\documentclass[10pt,twocolumn,letterpaper]{article}

\usepackage{cvpr}      
\definecolor{cvprblue}{rgb}{0.21,0.49,0.74}
\usepackage[pagebackref,breaklinks,colorlinks,allcolors=cvprblue]{hyperref}
\usepackage{colortbl}
\usepackage{multirow}

\def\paperID{****} 
\def\confName{CVPR}
\def\confYear{2026}

\title{ViT$^3$: Unlocking Test-Time Training in Vision}

\author{
  Dongchen~Han$^{1}$\hspace{2mm}
  Yining~Li$^{1}$\hspace{2mm}
  Tianyu~Li$^{1}$\hspace{2mm}
  Zixuan~Cao$^{1}$\hspace{2mm} \\
  Ziming~Wang$^{2}$\hspace{2mm}
  Jun~Song$^{2}$\hspace{2mm}
  Yu~Cheng$^{2}$\hspace{2mm}
  Bo~Zheng$^{2}$\hspace{2mm}
  Gao~Huang$^{1}$\thanks{Corresponding author.}\\
  {\small $^{1}$ Tsinghua University \hspace{6mm} $^2$ Alibaba Group}
}

\begin{document}
\maketitle

\begin{abstract}


Test-Time Training (TTT) has recently emerged as a promising direction for efficient sequence modeling. TTT reformulates attention operation as an online learning problem, constructing a compact inner model from key-value pairs at test time. This reformulation opens a rich and flexible design space while achieving linear computational complexity. However, crafting a powerful visual TTT design remains challenging: fundamental choices for the inner module and inner training lack comprehensive understanding and practical guidelines. To bridge this critical gap, in this paper, we present a systematic empirical study of TTT designs for visual sequence modeling. From a series of experiments and analyses, we distill six practical insights that establish design principles for effective visual TTT and illuminate paths for future improvement. These findings culminate in the Vision Test-Time Training (ViT$^3$) model, a pure TTT architecture that achieves linear complexity and parallelizable computation. We evaluate ViT$^3$ across diverse visual tasks, including image classification, image generation, object detection, and semantic segmentation. Results show that ViT$^3$ consistently matches or outperforms advanced linear-complexity models (e.g., Mamba and linear attention variants) and effectively narrows the gap to highly optimized vision Transformers. We hope this study and the ViT$^3$ baseline can facilitate future work on visual TTT models. 
Code: \href{https://github.com/LeapLabTHU/ViTTT}{github.com/LeapLabTHU/ViTTT}.

\end{abstract}    
\section{Introduction}
\label{sec:intro}

\begin{figure}[t]
    \centering
    \vskip -0.05in
    \includegraphics[width=0.79\linewidth]{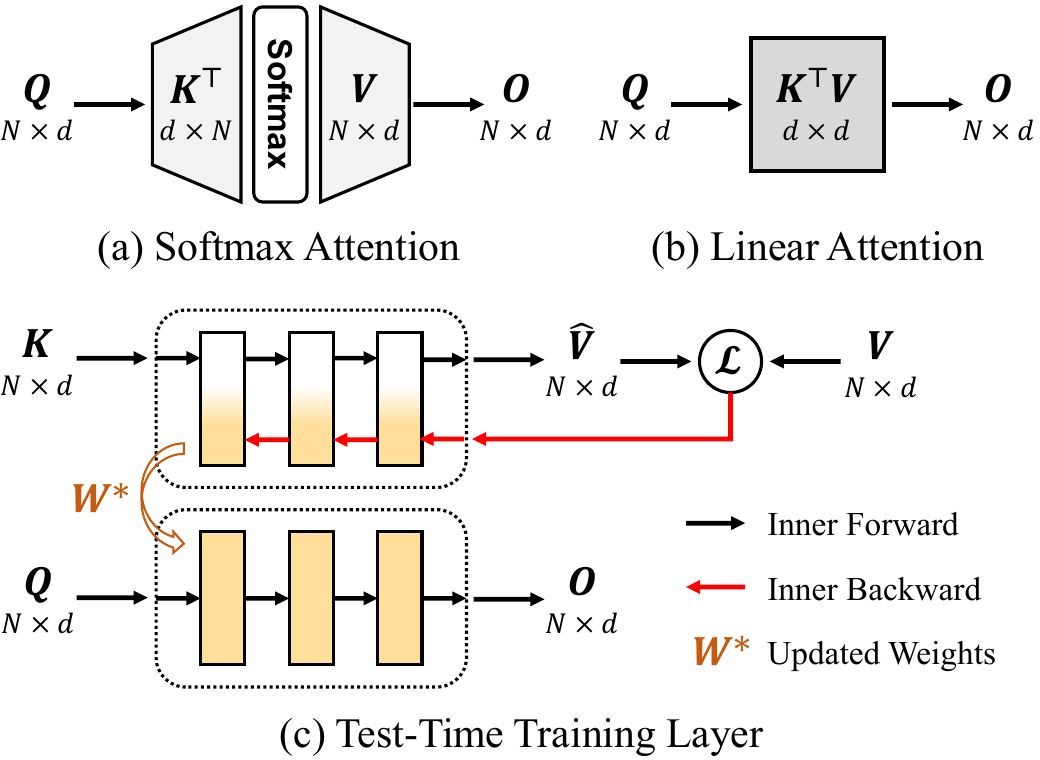}
    \vskip -0.05in
    \caption{Illustration of Softmax attention~\cite{attention}, linear attention~\cite{linear_attn}, and Test-Time Training (TTT) module~\cite{ttt}. (a) Softmax attention can be viewed as building a two-layer MLP that directly uses the uncompressed keys $K$ and values $V$, where the hidden width equals the sequence length $N$ and the nonlinearity is Softmax. While effective, this $N$-width MLP leads to $\mathcal{O}(N^2)$ costs when applied to the queries $Q\!\in\!\mathbb{R}^{N \times d}$. (b) Linear attention compresses $K,V\!\in\!\mathbb{R}^{N \times d}$ into a $d\!\times\!d$ linear layer weights using matrix multiplication $K^\top V$, yielding $\mathcal{O}(N)$ complexity. However, the $d\!\times\!d$ linear state is limited and the naive $K^\top V$ compression can discard important information, often resulting in inferior performance. (c) TTT generalizes the inner model to an \textit{arbitrary} module $\mathcal{F}_W(\cdot)\!:\mathbb{R}^d\!\rightarrow\!\mathbb{R}^d$ and compresses $K,V\!\in\!\mathbb{R}^{N \times d}$ into the module weights $W$ through a few self-supervised online training steps. Output is calculated with the updated weights $W^*$. When $\mathcal{F}_W(\cdot)$ is implemented as a linear-complexity module, e.g., a two-layer MLP with hidden dimension $4d$, TTT can deliver powerful $\mathcal{O}(N)$ sequence modeling. Please refer to \cref{sec:preliminaries} for details. }
    \label{fig:ttt}
    \vskip -0.2in
\end{figure}

Vision Transformers (ViT)~\cite{vit, deit} have become a cornerstone of modern computer vision, driving state-of-the-art results in image classification~\cite{vit,deit,swin}, generation~\cite{dit}, object detection~\cite{detr, deformabledetr}, and segmentation~\cite{segformer, gsva}. However, a fundamental limitation lies in the quadratic complexity $\mathcal{O}(N^2)$ of Softmax attention~\cite{attention} with respect to sequence length. This scaling bottleneck makes processing long visual sequences prohibitively expensive.

To break this quadratic barrier, the community has explored linear-complexity $\mathcal{O}(N)$ alternatives. A representative approach is linear attention~\cite{linear_attn}, which replaces Softmax with a linear kernel. This enables a reordering of computation from $(QK^T)V$ to $Q(K^T V)$, achieving the desired $\mathcal{O}(N)$ efficiency. 
Nonetheless, previous works~\cite{performer, cosformer, flatten} show that linear attention suffers from limited expressive power, making it impractical for many applications.

Test-Time Training (TTT) model~\cite{ttt}, a recent design, offers a new promising path forward.
Specifically, TTT reformulates the entire attention operation as an online learning process: in each context a compact inner model is trained or constructed from the key–value pairs and then applied to every query. As shown in \cref{fig:ttt}, from this viewpoint, Softmax attention can be interpreted as constructing a two-layer MLP with hidden dimension $N$ and Softmax activation, whereas linear attention corresponds to building a $d\!\times\!d$ linear layer via matrix multiplication $K^\top V$. In contrast to these \textit{fixed} designs, TTT allows the inner model to be an \textit{arbitrary} module $\mathcal{F}_W(\cdot)\!:\mathbb{R}^d\!\rightarrow\!\mathbb{R}^d$. It treats the key–value pairs as a ``mini-dataset'' and performs a few steps of self-supervised online training to learn (update) the parameters $W$ of this inner model at test time. This approach opens a far richer, more powerful linear-complexity design space, yielding impressive $\mathcal{O}(N)$ sequence modeling results in various scenarios~\cite{ttt_video, lact, ttt3r}.


However, the very flexibility of TTT introduces a critical bottleneck: a large and under-explored design space. Key choices regarding the inner training process and the inner model architecture lack systematic understanding in existing work~\cite{ttt, titans, lact}, which prevents the design of powerful visual TTT models. This gap motivates our central research question: \textit{What design principles can be established for building efficient yet expressive Test-Time Training module, and how to enable future improvement?}

To answer this question, we systematically explore the new design space of Test-Time Training models in vision. 
We focus on two fundamental aspects: inner training settings (loss function, learning rate, batch size, number of epochs), and inner model design (architecture and model size).
Through a series of experiments, we summarize our empirical observations into six practical insights:
1) loss functions $\mathcal{L}$ for which $\frac{\partial^2\mathcal{L}}{\partial \hat{V}\partial V}$ vanishes are not suitable for TTT;
2) a single epoch of full-batch (batch gradient) inner training works well for vision;
3) a relatively large inner learning rate (we use 1.0) is effective;
4) increasing inner model capacity consistently improves performance;
5) in current TTT settings, deep inner models suffer from optimization difficulties;
6) convolutional architectures are particularly appropriate as inner models for visual tasks.

Guided by these findings, we propose \textbf{Vision Test-Time Training (ViTTT, ViT$^3$)} model, a simple, parallelizable and pure TTT architecture tailored for visual sequence modeling.
We evaluate ViT$^3$ on a broad suite of tasks, including image classification, generation, object detection, and semantic segmentation.
ViT$^3$ consistently matches or outperforms advanced $\mathcal{O}(N)$ models, such as Mamba and linear attention variants, while effectively narrowing the performance gap to highly optimized $\mathcal{O}(N^2)$ vision Transformers.
These results confirm that the proposed ViT$^3$ offers impressive efficiency and capacity, demonstrating the potential of test-time training approaches on visual tasks.

Our main contributions and takeaways are as follows:

\begin{itemize}

    \item We present a systematic empirical study of Test-Time Training designs for vision, covering inner training regimes (loss function, learning rate, batch size, epochs) and inner model design (architecture and model size).
    
    \item We offer six practical insights for building effective yet efficient TTT module, providing detailed analyses of the TTT design space. Our analyses also reveal several valuable future research directions for TTT models.

    \item We build the Vision Test-Time Training (ViT$^3$) model, a simple TTT architecture that implements these insights. With $\mathcal{O}(N)$ complexity, ViT$^3$ achieves competitive results across image classification, image generation, object detection, and semantic segmentation, serving as a strong baseline for future research on visual TTT models.
\end{itemize}

\section{Related Work}
\label{sec:related_work}


\noindent
\textbf{Attention}~\cite{attention} \textbf{and vision Transformers}~\cite{vit, deit, cait, circulant}
have driven substantial progress in recent years. However, the quadratic computation complexity $\mathcal{O}(N^2)$ of self-attention leads to unmanageable cost when processing long visual sequences. To mitigate this, many works reduce the number of tokens involved in attention by introducing locality~\cite{swin, nat, cswin} or sparsity~\cite{pvt, dat, biformer}. An alternative family of approaches, linear attention methods~\cite{linear_attn, gated_linear_attn, mamba, deltanet}, redesign the attention paradigm to achieve global modeling with linear complexity $\mathcal{O}(N)$. With growing demand to process longer visual sequences, these linear-complexity models~\cite{nystromformer, flatten, vmamba, efficientvmamba} have attracted increasing interest. In this paper, we complement these lines of work by exploring TTT–based linear-complexity design.

\noindent
\textbf{Test-Time Training (TTT) model} 
~\cite{ttt} is a recently emerged approach for efficient sequence modeling, which reformulates attention operation as an online learning problem. This paradigm increases the flexibility of sequence models and has enabled scalable, linear-complexity designs across multiple domains~\cite{ttt_video, lact, atlas}. For instance, LaCT~\cite{lact} attains effective linear-time language modeling by applying large chunk test-time training. TTT layers allow pretrained diffusion Transformers that could only generate 3-second clips to produce coherent one-minute videos~\cite{ttt_video}. TTT3R~\cite{ttt3r} views 3D reconstruction as a test-time training process, achieving a $2\times$ improvement in global pose estimation. TTT is also closely related to fast weight programmers~\cite{fast_weight}, meta-learning~\cite{meta_learning, meta_learning_update}, Test-Time Scaling~\cite{tts, simple_tts} and recurrent depth~\cite{latent_reasoning, recurrent_memory_transformer}. Despite these promising results, the superiority of TTT methods remain locked by the under-explored design space (i.e., inner training and inner module settings). In this paper, we systematically evaluate TTT in vision, summarizing several insights for effective TTT models and potential future works.

\begin{figure}[t]
    \centering
    \includegraphics[width=0.8\linewidth]{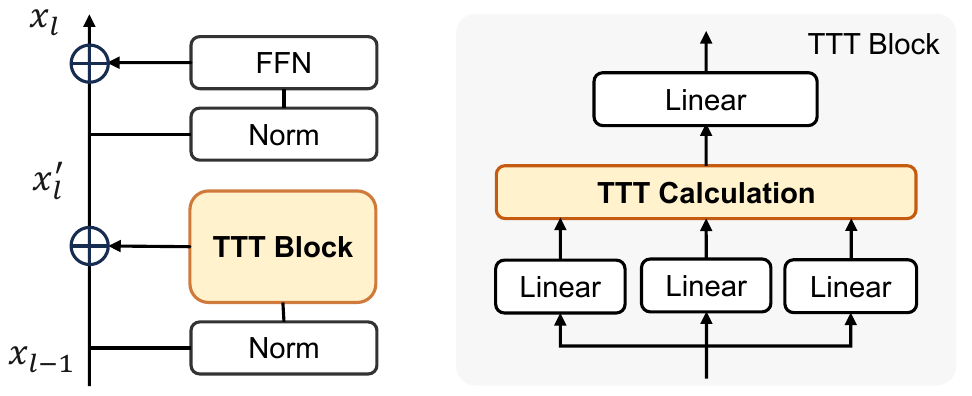}
    \vskip -0.1in
    \caption{Illustration of the TTT model building block. TTT shares the same macro architecture as Transformer. }
    \label{fig:ttt_block}
    \vskip -0.1in
\end{figure}

\section{Preliminaries} \label{sec:preliminaries}

\subsection{Attention Mechanism} \label{sec:soft_linear_attn}

\textbf{Softmax attention}~\cite{attention}, also referred to as dot-product attention, is the predominant attention mechanism used in modern Transformer architectures. Let $ x\!\in\!\mathbb{R}^{N \times C} $ denote an input sequence of $N$ tokens with dimension $C$. A single head of Softmax attention can be written as:
\begin{equation} \label{eq:softmax_attn1}
    \begin{split}
        O_i&=\sum_{j=1}^{N}\ \frac{\exp\left(Q_iK_j^\top\right)}{\sum_{j=1}^{N}\ \exp\left(Q_iK_j^\top\right)}V_j,\\
    \end{split}
\end{equation}
where $ Q\!=\!xW_Q$, $K\!=\!xW_K$, $V\!=\!xW_V$, $Q, K, V\!\in\!\mathbb{R}^{N \times d} $ represent queries, keys and values, and $ W_{Q/K/V}\!\in\!\mathbb{R}^{C \times d} $ are projection matrices. 
We omit the factor $\frac{1}{\sqrt{d}}$, as it can be achieved equivalently by scaling $Q,K$. To facilitate analysis, we reformulate \cref{eq:softmax_attn1} as:
\begin{equation} \label{eq:softmax_attn2}
    \begin{split}
        O&=\sigma(QK^\top)V\triangleq \sigma(QW_1)W_2=\mathrm{MLP}(Q),
    \end{split}
\end{equation}
where $\sigma(\cdot)$ is the row-wise Softmax operation. This expression highlights that Softmax attention can be viewed as a two-layer MLP of width $N$ acting on $Q$, with parameters $W_1\!=\!K^\top,W_2\!=\!V$, and Softmax serving as the activation. This $N$-width MLP leads to $\mathcal{O}(N^2)$ costs when evaluating $O\!=\!\mathrm{MLP}(Q)$, resulting in excessive computational costs in long-sequence modeling scenarios.

\noindent
\textbf{Linear attention}~\cite{linear_attn} is an alternative attention paradigm that addresses the quadratic cost by reducing complexity to $\mathcal{O}(N)$.
Specifically, linear attention replaces the non-linear Softmax with a linear kernel-based normalization:
\begin{equation} \label{eq:linear_attn1}
    \begin{split}
        O_i&=\!\sum_{j=1}^{N}\!\frac{Q_iK_j^{\top}}{\sum_{j=1}^{N}\!{Q_iK_j^{\top}}}V_j
        =\frac{Q_i\!\left(\sum_{j=1}^{N}\!{K_j^{\top}V_j}\right)}{Q_i\!\left(\ \sum_{j=1}^{N}\!{K_j^{\top}}\right)}, \\
    \end{split}
\end{equation}
where $ Q\!=\!\phi(xW_Q)$, $K\!=\!\phi(xW_K)$, $V\!=\!xW_V$, and $\phi(\cdot)$ is the kernel function.
This separate non-linearity on $Q$ and $K$ enables the rearrangement of the computation order from $(QK^{\top})V$ to $Q(K^{\top}V)$ based on the associative property of matrix multiplication, thus reducing complexity to $ \mathcal{O}(N) $. If we momentarily ignore the scalar normalization factor $Q_i\big(\sum_{j=1}^N\!K_j^\top\big)\!\in\!\mathbb{R}$, \cref{eq:linear_attn1} simplifies to:
\begin{equation} \label{eq:linear_attn2}
    \begin{split}
        O&=Q(K^\top V)\triangleq QW=\mathrm{FC}(Q).
    \end{split}
\end{equation}
Consequently, linear attention corresponds to compressing $K$ and $V$ into the $d\times d$ linear (FC) layer weights $W\!=\!K^\top V$.
While this approach enables $\mathcal{O}(N)$ efficiency, the limited linear state and naive $K^{\top}V$ compression tend to be inferior in practice~\cite{cosformer, flatten, inline}.

\subsection{Test-Time Training Model} \label{sec:ttt}

Test-Time Training (TTT)~\cite{ttt} reframes sequence modeling as an online learning problem and thereby generalizes linear attention into a unified, more powerful compression–and–adaptation framework. The key idea is to compress contextual key–value information $(K,V)$ into the weights of a compact neural network $\mathcal{F}_W\!:\mathbb{R}^d\rightarrow\mathbb{R}^d$, and then extract output features using queries.
The compression is achieved by a fast self-supervised online learning on the key-value ``dataset'' $\mathcal{D}=\{(K_i,V_i)|i=1,\cdots,N\}$:
\begin{equation} \label{eq:inner_training}
    \begin{split}
        \hat{V}_\mathcal{B}=\mathcal{F}_W(K_\mathcal{B}), \ \ 
        W\leftarrow W\! - \eta\! \cdot\! \frac{\partial\mathcal{L}(\hat{V}_\mathcal{B},V_\mathcal{B})}{\partial W},
    \end{split}
\end{equation}
where $K_\mathcal{B},V_\mathcal{B}\!\in\!\mathbb{R}^{B \times d}$ denote a mini-batch of size $B$ sampled from $\mathcal{D}$. $\mathcal{L}$ is a self-supervised reconstruction objective that encourages $\mathcal{F}_W(K_i)$ to predict the associated value $V_i$. The initial $W_0$ are implemented as learnable parameters of the outer network. After a few inner updates, the adapted parameters $W^*$ are used to produce the output $O\!=\!\mathcal{F}_{W^*}(Q)$. The approach is referred to as a Test-Time Training (TTT) layer because the inner module $\mathcal{F}_{W_0}$ is briefly trained (adapted) per input sequence at test time. Importantly, the inner adaptation procedure of \cref{eq:inner_training} is differentiable and is optimized jointly with the outer network during training (i.e., the inner updates are unrolled and learned end-to-end), so models with TTT layers remain fully end-to-end trainable. We refer to $\mathcal{F}_W$ and \cref{eq:inner_training} as \textit{inner module} and \textit{inner training (inner loop)}, and refer to the entire network and its training on real data as \textit{outer model} and \textit{outer training (outer loop)}.

The per-sequence computation and memory cost of a TTT layer scale with the cost of the inner module. Therefore, when the inner module is implemented using a linear-complexity architecture, e.g., a two-layer MLP with ratio 4.0, the TTT layer inherits $\mathcal{O}(N)$ time and memory complexity while benefiting from the richer, learning-based compression and non-linear expressive states.

\section{Exploring Test-Time Training Designs}
\label{sec:method}

In this section, we conduct a systematic empirical study of Test-Time Training designs in vision. We employ the classical vision Transformer architecture DeiT-S~\cite{deit} and substitute the attention blocks with TTT layers to create our baseline. Each inner training (loss function, learning rate, batch size and epochs) and model (architecture and size) designs are introduced \textit{separately} to the baseline model to assess their impact. 
Experiments are conducted on ImageNet-1K with the standard 300 epoch training settings.

\subsection{Inner Training Configuration}
\label{sec:inner_training}

We first implement the inner model as a two-layer MLP (SiLU activation, ratio 1.0) to study the inner training.

\vskip 0.1cm
\noindent
\textbf{Insight 1: inner loss functions $\mathcal{L}$ for which the mixed second derivative $\frac{\partial^2\mathcal{L}}{\partial \hat{V}\partial V}$ vanishes are not suitable for TTT.} 
As discussed in \cref{sec:preliminaries}, the inner training process of TTT layer is optimized jointly with the outer network. Concretely, the value projection matrix $ W_V\!\in\!\mathbb{R}^{C \times d} $ obtains outer-loop gradients through backpropagation applied to the inner update term $G\!=\!\frac{\partial\mathcal{L}(\hat{V}_\mathcal{B},V_\mathcal{B})}{\partial W}$ in \cref{eq:inner_training}. In other words, we differentiate through the inner training steps and taking gradients of gradients, which is a commonly studied operation in meta-learning~\cite{meta_learning}. Note that
\begin{equation}
    \frac{\partial G}{\partial W_V}\!=\!\frac{\partial^2\mathcal{L}(\hat{V}_\mathcal{B},V_\mathcal{B})}{\partial W\partial W_V}\!=\!\frac{\partial\hat{V}_\mathcal{B}}{\partial W}\cdot\frac{\partial^2\mathcal{L}(\hat{V}_\mathcal{B},V_\mathcal{B})}{\partial \hat{V}_\mathcal{B}\partial V_\mathcal{B}}\cdot\frac{\partial V_\mathcal{B}}{\partial W_V}.
\end{equation}
If the mixed derivative $\frac{\partial^2\mathcal{L}}{\partial \hat{V}\partial V}$ is (near) zero, the outer-loop gradient signal to $W_V$ vanishes after backpropagation through the inner step, which undermines learning and degrades performance. Our experiments in \cref{tab:loss_function} support this analysis: MAE (L1) loss — whose derivative is a sign function and whose mixed second derivative is zero almost everywhere — yields the worst accuracy. Smooth L1 loss also performs poorly because its mixed derivative vanishes in certain regions, whereas the three other losses which do not exhibit this issue achieve approximately 79.0\% accuracy.

\noindent
\textbf{Remark 1.} Follow prior TTT work~\cite{ttt, lact}, we focus on loss functions that encourages $\mathcal{F}_W(K_i)$ to predict its corresponding value $V_i$. This training target is implicitly satisfied by the widely used Softmax attention~\cite{attention}. As analyzed in \cref{sec:preliminaries}, Softmax attention can be viewed as constructing an inner two-layer MLP $\mathcal{F}(\cdot)$ using $K$ and $V$. For an input $K_i$, the output of this inner MLP is $\mathcal{F}(K_i)\!=\!\sigma(K_iK^\top)V$, where $\sigma(\cdot)$ denotes the row-wise Softmax. Under common conditions (e.g., sufficiently distinct key vectors and appropriate temperature scaling), the self-similarity $K_i K_i^\top$ is typically the largest value in $K_i K^\top$, so $\sigma(K_i K^\top)$ concentrates most score on the $i$-th position and becomes approximately one-hot. Therefore, $\mathcal{F}(K_i)\!=\!\sigma(K_iK^\top)V\!\approx\!V_i$, which is exactly the training target of TTT.

\begin{table}[t]
    \centering
    \footnotesize
    \setlength{\tabcolsep}{1.0mm}{
    \renewcommand\arraystretch{0.9}
    \begin{tabular}{l|c c|c c}
        \toprule
        \textbf{Loss Function}  &  \textbf{\#Params} & \textbf{FLOPs}  & \textbf{FPS}    & \textbf{Top-1}\\
        \midrule
        Dot Product Loss
        & 23.5M     & 4.58G     & 1315 & 78.9\\
        MSE (L2) Loss
        & 23.5M     & 4.63G     & 1296 & 79.2\\
        RMSE Loss
        & 23.5M     & 4.63G     & 1269 & 78.8\\
        MAE (L1) Loss
        & 23.5M     & 4.63G     & 1292 & 76.5\\
        Smooth L1 loss
        & 23.5M     & 4.63G     & 1292 & 78.1\\
        \bottomrule
    \end{tabular}}
    \vskip -0.2cm
    \caption{Results of different inner training loss functions. Please refer to the Appendix for detailed formulas of each loss function.}
    \label{tab:loss_function}
    \vskip -0.2cm
\end{table}

\noindent
\textbf{Insight 2: a single epoch of full-batch (batch gradient) inner training works well for vision.}
We first study the effect of the inner-training batch size, presenting the results in \cref{tab:batch_size_epoch}. Here $B\!=\!N$ denotes full-batch gradient descent using all $N$ key–value pairs as a single inner batch, whereas $B\!=\!\frac{N}{2},\frac{N}{3},\frac{N}{4}$ refer to mini-batch gradient descent, partitioning the dataset into $2, 3, 4$ sequential mini-batches and conducting $2, 3, 4$ inner updates per epoch. In contrast to prior TTT results on language modeling that reported gains from smaller mini-batches~\cite{ttt}, we find that $B\!=\!N$ leads to the best outcome in our vision experiments. We attribute this discrepancy to the difference between causal and non-causal data modalities. Specifically, sequential mini-batch gradient descent imposes a causal bias: (1) updates from earlier mini-batches change the inner-module weights, thus influencing the gradients of later batches; (2) updates from later mini-batches can overwrite those produced earlier. This causal dependency is ideally suited for causal data like language, but could be suboptimal for vision~\cite{demystify_mamba, mambaout}.

Multiple epochs of full-batch inner-training improve accuracy. However, they notably reduce model throughput and can introduce training instability.

\noindent
\textbf{Remark 2.} 
Existing linear sequence-modeling approaches for vision (e.g., linear attention~\cite{flatten, demystify_mamba} and Mamba~\cite{vmamba, mambatree}) can be categorized into two broad classes: parallel and sequential. Parallel designs~\cite{flatten,demystify_mamba} model the entire input sequence in a non-causal (global) manner,  which is similar to the single-epoch, full-batch inner update in our TTT setup. In contrast, sequential models~\cite{vmamba, mambatree} process inputs along carefully designed scan paths, capturing richer and more expressive spatial dependencies. Therefore, while naive sequential mini-batch exhibits subpar performance in our experiments, designing mini-batch inner-training algorithms tailored for vision remains a promising future work.

\begin{table}[t]
    \centering
    \footnotesize
    \setlength{\tabcolsep}{1.5mm}{
    \renewcommand\arraystretch{0.9}
    \begin{tabular}{c c|c c |c c}
        \toprule
        \textbf{Epoch} & \textbf{Batch Size}    &  \textbf{\#Params} & \textbf{FLOPs}  & \textbf{FPS}    & \textbf{Top-1}\\
        \midrule
        1 & $N$
        & 23.5M     & 4.58G     & 1315 & 78.9\\
        1 & $N/2$
        & 23.5M     & 4.58G     & 1201 & 78.6\\
        1 & $N/3$
        & 23.5M     & 4.58G     & 1131 & 78.3\\
        1 & $N/4$
        & 23.5M     & 4.58G     & 1101 & 78.1\\
        2 & $N$
        & 23.5M     & 4.81G     & 971  & 79.1\\
        3 & $N$
        & 23.5M     & 5.04G     & 787  & 79.2\\
        4 & $N$
        & 23.5M     & 5.27G     & 659  & 57.0*\\
        \bottomrule
    \end{tabular}}
    \vskip -0.2cm
    \caption{Results of various batch sizes and epochs. * refers to the best accuracy before divergence during training.}
    \label{tab:batch_size_epoch}
    \vskip -0.2cm
\end{table}

\begin{table}[t]
    \centering
    \footnotesize
    \setlength{\tabcolsep}{0.5mm}{
    \renewcommand\arraystretch{0.9}
    \begin{tabular}{l|c c c c c c c c}
        \textbf{Learning Rate} & 0.1 & 0.2 & 0.5 & 1.0 & 2.0 & 5.0 & 10.0 & Dynamic\\
        \midrule
        \textbf{Top-1} & 77.5 & 78.1 & 78.7 & 78.9 & 78.9 & 76.7* & 76.9* & 78.7\\
    \end{tabular}}
    \vskip -0.2cm
    \caption{Results of different inner learning rates. * refers to the best accuracy before divergence during training.}
    \label{tab:learning_rate}
    \vskip -0.3cm
\end{table}

\vskip 0.1cm
\noindent
\textbf{Insight 3: a relatively large inner learning rate of 1.0 is effective.}
\cref{tab:learning_rate} reports the results of different inner learning rates, ranging from 0.1 to 10. Small inner learning rates produce insufficient updates to the inner model weights $W$, while excessively large ones can result in training instability. We observe that $\eta\!=\!1.0$ yields meaningful inner updates while preserving stability of the outer optimization. We additionally evaluate an input-dependent, token-wise dynamic rate $\eta_i\!=\!\eta\!\cdot\!\mathrm{Sigmoid}(x_iW_\eta),\eta\!=\!1.0$ as proposed in prior work~\cite{ttt, lact}. In our vision experiments this scheme is generally less effective.

\noindent
\textbf{Remark 3.}
In some special cases, the inner learning rate can be absorbed into the scaling of $K$ and $V$. For example, when the inner model is a linear layer $\mathcal{F}_W(x)=xW,W\in\mathbb{R}^{d\times d}$ and the inner loss is mean squared error (MSE), the gradient-based update term can be written as:
\begin{equation}
    \eta\cdot\frac{\partial\mathcal{L}(\hat{V}, V)}{\partial W}=\eta \cdot K^\top (KW\!-\!V) =\tilde{K}^\top (\tilde{K}W\!-\!\tilde{V}),
\end{equation}
where $\tilde{K}\!=\!\sqrt{\eta}K,\tilde{V}\!=\!\sqrt{\eta}V$. 
This shows that scaling $K$ and $V$ is mathematically equivalent to changing $\eta$. Nevertheless, $\eta$ remains a crucial practical hyperparameter because such rescaling could be hard to learn or incompatible with other model components (e.g., initialized parameter scales, or normalization layers). The impact of inner learn rate is fully demonstrated by our results in \cref{tab:learning_rate}. Similarly, the famous $\frac{1}{\sqrt{d}}$ scaling in Softmax attention can be absorbed into $Q$ and $K$, but is well known to be critical in practice~\cite{attention}.

\subsection{Inner Model Design}
\label{sec:inner_model}

To study the inner model designs, we hold the inner training configuration as follows: dot product loss, one epoch of full-batch gradient descent, and learning rate 1.0.

\begin{table}[t]
    \centering
    \footnotesize
    \setlength{\tabcolsep}{1.5mm}{
    \renewcommand\arraystretch{0.9}
    \begin{tabular}{l|c c |c c}
        \toprule
        \textbf{Inner Model}  &  \textbf{\#Params} & \textbf{FLOPs}  & \textbf{FPS}    & \textbf{Top-1}\\
        \midrule
        $\mathrm{MLP}(x),\ r1,l2$
        & 23.5M     & 4.58G     & 1315 & 78.9\\
        $\mathrm{MLP}(x),\ r2,l2$
        & 24.1M     & 4.92G     & 1119 & 79.2\\
        $\mathrm{MLP}(x),\ r3,l2$
        & 24.7M     & 5.27G     & 938  & 79.5\\
        $\mathrm{MLP}(x),\ r4,l2$
        & 25.2M     & 5.62G     & 836  & 79.6\\
        
        \midrule
        $\mathrm{FC}(x)$
        & 23.2M     & 4.34G     & 1708 & 79.1\\
        $\mathrm{MLP}(x),\ r1,l2$
        & 23.5M     & 4.58G     & 1315 & 78.9\\
        $\mathrm{MLP}(x),\ r1,l3$
        & 23.8M     & 4.81G     & 1086 & 77.5\\
        $\mathrm{SiLU}(\mathrm{FC}(x))$
        & 23.2M     & 4.40G     & 1456 & 79.4\\
        $\mathrm{SwiGLU}(x)$
        & 23.8M     & 4.75G     & 1103 & 79.0\\
        $\mathrm{FC}(x)\odot\mathrm{SiLU}(\mathrm{FC}(x))$
        & 23.5M     & 4.58G     & 1194 & 79.7\\

        \midrule
        $\mathrm{Conv}(x)$
        & 25.5M     & 5.27G     & 979  & 79.9\\
        $\mathrm{DWConv}(x)$
        & 22.9M     & 4.25G     & 1366 & 80.1\\
        \bottomrule
    \end{tabular}}
    \vskip -0.3cm
    \caption{Results of different inner model designs. The $r$ and $l$ denote width ratio and layer-wise depth of a MLP. For example, $r3,l2$ refers to a 2-layer MLP with hidden dimension $3d$, where $d$ is the input and output dimension of an inner model.}
    \label{tab:inner_model}
    \vskip -0.2cm
\end{table}

\vskip 0.1cm
\noindent
\textbf{Insight 4: increasing inner model capacity consistently improves performance.}
To validate this, we instantiate the inner model $\mathcal{F}_W(\cdot)\!:\mathbb{R}^d\!\rightarrow\!\mathbb{R}^d$ as a two-layer MLP with SiLU activation and progressively enlarge its hidden dimension from $d$ to $4d$. \cref{tab:inner_model} reports that the accuracy increases consistently as inner model capacity grows. This demonstrates a key advantage of the TTT paradigm over previous linear attention methods~\cite{linear_attn, cosformer, inline}: rather than restricting the inner model to a linear $d\!\times\!d$ mapping, TTT allows for a more complex nonlinear module as the inner model, and consistently benefit from its capacity. 

\noindent
\textbf{Remark 4.} Increasing inner model size incurs more computation than enlarging an outer model. At test time, an outer module $\mathcal{F}_W(\cdot)\!:\mathbb{R}^d\!\rightarrow\!\mathbb{R}^d$ requires only a single forward pass $y\!=\!\mathcal{F}_W(x)$. In contrast, an inner module needs to perform: (i) a forward pass on keys $\hat{V}\!=\!\mathcal{F}_W(K)$, (ii) the backward pass of loss $\mathcal{L}$, and (iii) a forward pass on queries $O\!=\!\mathcal{F}_W(Q)$. A backward pass typically costs twice the FLOPs of a forward pass, because backpropagation computes gradients of both parameter and input; for example, for the forward $Y\!=\!XW$ of a linear layer, we compute $\frac{\partial\mathcal{L}}{\partial W} = X^\top \frac{\partial\mathcal{L}}{\partial Y}$ and $\frac{\partial\mathcal{L}}{\partial X} = \frac{\partial\mathcal{L}}{\partial Y} W^\top$ (to propagate to previous layers). Hence, one inner training epoch consumes approximately $1+2+1\!=\!4$ forward-equivalent FLOPs, i.e., about $4\times$ the compute of an outer module with the same architecture. Therefore, while simply scaling inner model demonstrates promising results, we believe designing lightweight and expressive inner model is an important topic.

\begin{figure}[t]
    \centering
    \includegraphics[width=0.9\linewidth]{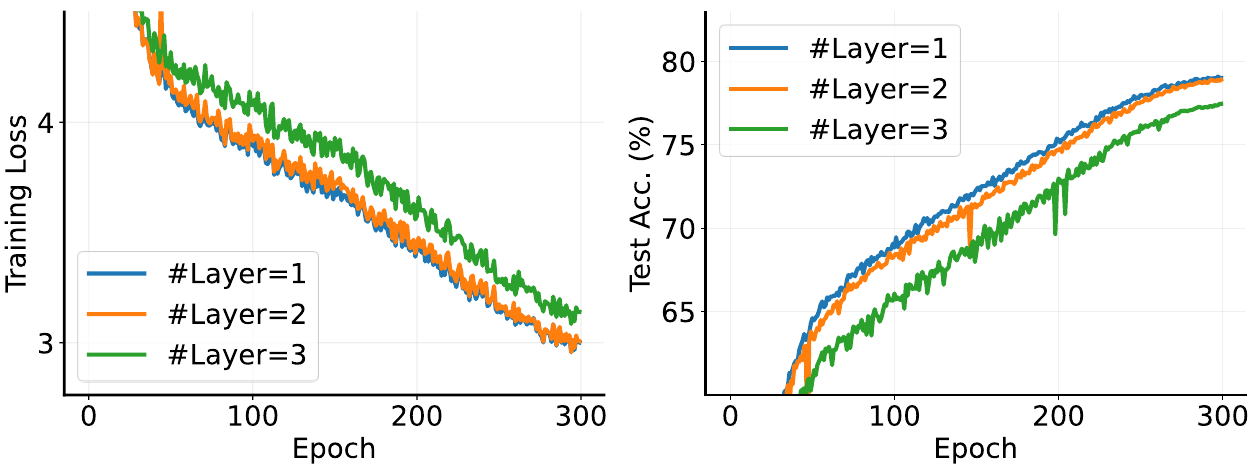}
    \vskip -0.1in
    \caption{Results of TTT models with inner modules of $1,2,3$ layers (FC, two-layer and three-layer MLP). Deeper inner models lead to higher training loss, and thus lower test accuracy.}
    \label{fig:loss_acc}
    \vskip -0.2in
\end{figure}

\begin{table*}[t]

    \centering
    \begin{minipage}[t]{0.495\linewidth}
        \centering
        \footnotesize
        \setlength{\tabcolsep}{1.8mm}{
        \renewcommand\arraystretch{0.9}
        \begin{tabular}{l|c|c c|c}
            \toprule
            Method
            & Type          & \#Params  & FLOPs     & Top-1 \\
            
            \midrule
            ConvNeXt-T~\cite{convnext}
            & ConvNet       & 29M       & 4.5G      & 82.1 \\
            InternImage-T~\cite{internimage}
            & ConvNet       & 30M       & 5.0G      & 83.5 \\
            BiFormer-S~\cite{biformer}
            & Transformer   & 26M       & 4.5G      & 83.8 \\
            TransNeXt-T~\cite{transnext}
            & Transformer   & 28M       & 5.7G      & 84.0 \\
            RMT-S~\cite{rmt}
            & Transformer   & 27M       & 4.5G      & 84.1 \\
            FasterViT-1$^\ddag$~\cite{fastervit}
            & Transformer   & 53M       & 5.3G      & 83.2 \\
            VMamba-T~\cite{vmamba} 
            & Mamba         & 31M       & 4.9G      & 82.5 \\
            LocalVMamba-T~\cite{localmamba} 
            & Mamba         & 26M       & 5.7G      & 82.7 \\
            SOFT-S++~\cite{soft_plus}
            & Linear        & 27M       & 4.5G      & 82.6 \\
            VVT-S~\cite{vvt}
            & Linear        & 26M       & 5.6G      & 82.7 \\
            MILA-T$^\ddag$~\cite{demystify_mamba}
            & Linear        & 25M       & 4.2G      & 83.5 \\
            \rowcolor{lightgray!20} H-ViT$^3$-T
            & TTT           & 29M       & 4.9G      & 83.5 \\
            \rowcolor{lightgray!20} H-ViT$^3$-T$^\ddag$
            & TTT           & 29M       & 4.9G      & 84.0 \\

            \midrule
            InceptionNeXt-B~\cite{inceptionnext}
            & ConvNet       & 87M       & 14.9G     & 84.0 \\
            InternImage-B~\cite{internimage}
            & ConvNet       & 97M       & 16.0G     & 84.9 \\
            CSwin-B~\cite{cswin}
            & Transformer   & 78M       & 15.0G     & 84.2 \\
            iFormer-L~\cite{iformer}
            & Transformer   & 87M       & 14.0G     & 84.8 \\
            TransNeXt-B~\cite{transnext}
            & Transformer   & 90M       & 18.4G     & 84.8 \\
            FasterViT-3$^\ddag$~\cite{fastervit}
            & Transformer   & 160M      & 18.2G     & 84.9 \\

            \bottomrule
        \end{tabular}}
    \end{minipage}
    \begin{minipage}[t]{0.495\linewidth}
        \centering
        \footnotesize
        \setlength{\tabcolsep}{1.8mm}{
        \renewcommand\arraystretch{0.89}
        \begin{tabular}{l|c|c c|c}
            \toprule
            Method
            & Type          & \#Params  & FLOPs     & Top-1 \\

            \midrule
            ConvNeXt-S~\cite{convnext}
            & ConvNet       & 50M       & 8.7G      & 83.1 \\
            InternImage-S~\cite{internimage}
            & ConvNet       & 50M       & 8.0G      & 84.2 \\
            BiFormer-B~\cite{biformer}
            & Transformer   & 57M       & 9.8G      & 84.3 \\
            TransNeXt-S~\cite{transnext}
            & Transformer   & 50M       & 10.3G     & 84.7 \\
            RMT-B~\cite{rmt}
            & Transformer   & 54M       & 9.7G      & 85.0 \\
            FasterViT-2$^\ddag$~\cite{fastervit}
            & Transformer   & 76M       & 8.7G      & 84.2 \\
            VMamba-S~\cite{vmamba} 
            & Mamba         & 50M       & 8.7G      & 83.6 \\
            LocalVMamba-S~\cite{localmamba} 
            & Mamba         & 50M       & 11.4G     & 83.7 \\
            SOFT-M++~\cite{soft_plus}
            & Linear        & 48M       & 8.7G      & 83.7 \\
            VVT-M~\cite{vvt}
            & Linear        & 48M       & 9.4G      & 83.8 \\
            MILA-S$^\ddag$~\cite{demystify_mamba}
            & Linear        & 43M       & 7.3G      & 84.4 \\
            \rowcolor{lightgray!20} H-ViT$^3$-S
            & TTT           & 54M       & 8.8G      & 84.4 \\
            \rowcolor{lightgray!20} H-ViT$^3$-S$^\ddag$
            & TTT           & 54M       & 8.8G      & 84.9 \\

            \midrule
            Mamba2D-B~\cite{mamband} 
            & Mamba         & 94M       & $-$       & 83.0 \\
            VMamba-B~\cite{vmamba} 
            & Mamba         & 89M       & 15.4G     & 83.9 \\
            SOFT-L++~\cite{soft_plus}
            & Linear        & 85M       & 15.4G     & 84.1 \\
            MILA-B$^\ddag$~\cite{demystify_mamba}
            & Linear        & 96M       & 16.2G     & 85.3 \\
            \rowcolor{lightgray!20} H-ViT$^3$-B
            & TTT           & 94M       & 16.7G     & 84.9 \\
            \rowcolor{lightgray!20} H-ViT$^3$-B$^\ddag$
            & TTT           & 94M       & 16.7G     & 85.5 \\
            
            \bottomrule
        \end{tabular}}
    \end{minipage}
    \vskip -0.15cm
    \caption{Comparison with hierarchical architectures on ImageNet-1K. We focus on representative ConvNet, Transformer, Mamba, and Linear attention methods. \ddag\ indicates the model is trained with MESA~\cite{mesa}, a strategy that can alleviate overfitting at little cost.}
    \label{tab:cls1}
    \vskip -0.4cm
\end{table*}

\begin{table}[t]
    \centering
    \footnotesize
    \setlength{\tabcolsep}{1.5mm}{
    \renewcommand\arraystretch{0.9}
    \begin{tabular}{l|c c |c c}
        \toprule
        \textbf{Inner Model}  &  \textbf{\#Params} & \textbf{FLOPs}  & \textbf{FPS}    & \textbf{Top-1}\\
        \midrule
        $\mathrm{SiLU}(xW_1)W_2\!+\!x$
        & 23.5M     & 4.58G     & 1294 & 78.8\\
        $\mathrm{SiLU}(xW_1)(W_2\!+\!I)$
        & 23.5M     & 4.58G     & 1294 & 79.1\\
        $\mathrm{SiLU}(xW_1)W_2$, $W_2$ init as $I$
        & 23.5M     & 4.58G     & 1315 & 79.0\\
        \bottomrule
    \end{tabular}}
    \vskip -0.2cm
    \caption{Results of applying residual connections and weight initialization strategies to the MLP inner model. Despite achieving some improvements, these methods still underperform the constrained design $\mathcal{F}_W(x)\!=\!\mathrm{SiLU}(\mathrm{FC}(x))$, i.e., $\mathrm{SiLU}(xW_1)$.}
    \label{tab:residual_init}
    \vskip -0.5cm
\end{table}

\vskip 0.1cm
\noindent
\textbf{Insight 5: in current TTT settings, deep inner models suffer from optimization difficulties.}
Scaling inner model width leads to consistent gains, as suggested by Insight 4. A common alternative is to scale network depth~\cite{resnet, deepnet, cait}. We evaluate three inner models of different depths: a single $d\!\times\!d$ linear (FC) layer, a two-layer MLP, and a three-layer MLP. For both MLP variants, the hidden dimensions are set to $d$, matching the input and output dimensionality. \cref{tab:inner_model} shows that the final test accuracy does not scale with inner model depth, but instead demonstrates a clear degradation. This finding is counterintuitive, because deeper inner modules contain more parameters and therefore greater capacity.

We hypothesize that increasing the depth of the inner module exacerbates optimization difficulties, leading to diminishing returns. To validate this hypothesis, \cref{fig:loss_acc} plots training loss and test accuracy for the three architectures. Models with deeper inner modules exhibit higher training loss, and thus lower test accuracy. In other words, while deeper inner modules are theoretically more expressive and easier to fit the training data, they practically deliver lower performance due to underfitting. This disparity between theory and practice suggests an optimization bottleneck. We further corroborate this claim using a constrained inner-model $\mathcal{F}_W(x)\!=\!\mathrm{SiLU}(\mathrm{FC}(x))$, which can be viewed as a two-layer MLP whose output linear layer is fixed to the identity. Replacing the full two-layer MLP with this constrained design increases accuracy from 78.9\% to 79.4\%. A similar effect is also observed with SwiGLU~\cite{glu}: removing the output layer (i.e., a SwiGLU with identity output layer) raises the accuracy from 79.0\% to 79.7\%. These constrained special cases achieve much better results than original designs, validating the optimization problem. Together, these results fully support our hypothesis that deeper inner modules are harder to optimize and can therefore underperform their shallower counterparts in current settings.

\noindent
\textbf{Remark 5.}
The optimization difficulties of deep inner models can be attributed to two complementary aspects:
\begin{itemize}
    \item Outer-loop problem. The inner model initialization $W_0$ of $\mathcal{F}_W(\cdot)$ (i.e., initial weights for \cref{eq:inner_training}) is learned end-to-end as part of the full network parameters. For deep inner modules, this initialization becomes difficult to optimize during outer-loop (end-to-end) training.

    \item Inner-loop problem. Increasing inner depth causes exploding or vanishing inner-loop gradients $\frac{\partial\mathcal{L}}{\partial W}$, which hinder effective compression of the $K, V$ contest.
\end{itemize}
We empirically find that standard residual connections~\cite{resnet} and naive initialization schemes provide limited mitigation for these issues (see \cref{tab:residual_init}). Notably, theoretical works~\cite{shallow_vs_deep, number_of_linear, expressive_power, exponential_expressivity, depth_power} suggest that deeper neural networks offer exponentially greater capability, which is a key factor behind the success of modern deep learning models. Therefore, addressing the optimization challenges and enabling deep inner models within the TTT framework represents a fundamental and promising research direction.

\begin{table}[t]

    \centering
    \footnotesize
    \setlength{\tabcolsep}{1.2mm}{
    \renewcommand\arraystretch{0.85}
        \begin{tabular}{l|c|c c|c}
            \toprule
            Method
            & Type          & \#Params  & FLOPs     & Top-1 \\ 

            \midrule
            DeiT-T~\cite{deit}
            & Transformer   & 6M        & 1.2G      & 72.2 \\
            Vim-T~\cite{vim}
            & Mamba         & 7M        & 1.5G      & 76.1 \\
            Agent-DeiT-T~\cite{agent_attention}
            & Linear        & 6M        & 1.2G      & 74.9 \\
            \rowcolor{lightgray!20} ViT$^3$-T
            & TTT           & 6M        & 1.2G      & 76.5 \\

            \midrule
            ConvNeXt-S (iso.)~\cite{convnext}
            & ConvNet       & 22M       & 4.3G      & 79.7 \\
            DeiT-S~\cite{deit}
            & Transformer   & 22M       & 4.6G      & 79.8 \\
            Vim-S~\cite{vim}
            & Mamba         & 26M       & 5.1G      & 80.3 \\
            Agent-DeiT-S~\cite{agent_attention}
            & Linear        & 23M       & 4.4G      & 80.5 \\
            \rowcolor{lightgray!20} ViT$^3$-S
            & TTT           & 24M       & 4.8G      & 81.6 \\

            \midrule
            ConvNeXt-B (iso.)~\cite{convnext}
            & ConvNet       & 87M       & 16.9G     & 82.0 \\
            DeiT-B~\cite{deit}
            & Transformer   & 87M       & 17.6G     & 81.8 \\
            \rowcolor{lightgray!20} ViT$^3$-B
            & TTT           & 90M       & 18.0G     & 82.6 \\

            \bottomrule
        \end{tabular}}
    \vskip -0.2cm
    \caption{Comparison with non-hierarchical designs on ImageNet.}
    \label{tab:cls2}
    \vskip -0.6cm
\end{table}

\begin{table*}[t]

    \centering
    \begin{minipage}[t]{0.495\linewidth}
    \centering
    \footnotesize
    \setlength{\tabcolsep}{0.6mm}{
    \renewcommand\arraystretch{0.9}
    \begin{tabular}{lcccccccc}
        \toprule
        Method & Type & FLOPs & AP$^b$ & AP$^b_\text{50}$ & AP$^b_\text{75}$ & AP$^m$ & AP$^m_\text{50}$ & AP$^m_\text{75}$ \\
    
        \midrule
        \multicolumn{9}{l}{\textbf{\textit{Mask R-CNN $\ \ 1\times\!$ Schedule}}} \\
        \midrule
    
        InternImage-T~\cite{internimage}
        & C & 270G & 47.2 & 69.0 & 52.1 & 42.5 & 66.1 & 45.8 \\
        CSWin-T~\cite{cswin}
        & T & 279G & 46.7 & 68.6 & 51.3 & 42.2 & 65.6 & 45.4 \\
        FocalNet-T~\cite{focal}
        & T & 268G & 46.1 & 68.2 & 50.6 & 41.5 & 65.1 & 44.5 \\
        Vmamba-T~\cite{vmamba}
        & M & 271G & 47.3 & 69.3 & 52.0 & 42.7 & 66.4 & 45.9 \\
        SOFT-T++~\cite{soft_plus}
        & L & -    & 43.8 & 66.0 & 47.5 & 40.1 & 63.0 & 43.0 \\
        MILA-T~\cite{demystify_mamba}
        & L & 255G & 46.8 & 69.5 & 51.5 & 42.1 & 66.4 & 45.0 \\
        \rowcolor{lightgray!20} H-ViT$^3$-T
        &   & 271G & 47.3 & 69.8 & 52.3 & 42.8 & 66.8 & 46.2 \\

        \midrule
        InternImage-S~\cite{internimage}
        & C & 340G & 47.8 & 69.8 & 52.8 & 43.3 & 67.1 & 46.7 \\
        CSWin-S~\cite{cswin}
        & T & 342G & 47.9 & 70.1 & 52.6 & 43.2 & 67.1 & 46.2 \\
        TransNeXt-T~\cite{transnext}
        & T & 356G & 49.9 & 71.5 & 54.9 & 44.6 & 68.6 & 48.1 \\
        Vmamba-S~\cite{vmamba}
        & M & 357G & 48.7 & 70.0 & 53.4 & 43.7 & 67.3 & 47.0 \\
        SOFT-S++~\cite{soft_plus}
        & L & -    & 46.6 & 67.8 & 51.2 & 42.0 & 64.8 & 45.2 \\
        \rowcolor{lightgray!20} H-ViT$^3$-S
        &   & 349G & 49.1 & 71.5 & 53.7 & 44.1 & 68.4 & 47.4 \\

        \midrule
        InternImage-B~\cite{internimage}
        & C & 501G & 48.8 & 70.9 & 54.0 & 44.0 & 67.8 & 47.4 \\
        CSWin-B~\cite{cswin}
        & T & 526G & 48.7 & 70.4 & 53.9 & 43.9 & 67.8 & 47.3 \\
        TransNeXt-S~\cite{transnext}
        & T & 516G & 51.1 & 72.6 & 56.2 & 45.5 & 69.8 & 49.1 \\
        VMamba-B~\cite{vmamba}
        & M & 485G & 49.2 & 70.9 & 53.9 & 43.9 & 67.7 & 47.6 \\
        SOFT-B++~\cite{soft_plus}
        & L & -    & 47.0 & 68.3 & 51.7 & 42.2 & 65.2 & 45.4 \\
        \rowcolor{lightgray!20} H-ViT$^3$-B
        &   & 510G & 50.0 & 71.8 & 55.0 & 44.6 & 69.0 & 47.7 \\
        
        \bottomrule
    \end{tabular}}
    \end{minipage}
    \begin{minipage}[t]{0.495\linewidth}
    \centering
    \footnotesize
    \setlength{\tabcolsep}{0.6mm}{
    \renewcommand\arraystretch{0.9}
    \begin{tabular}{lcccccccc}
        \toprule
        Method & Type & FLOPs & AP$^b$ & AP$^b_\text{50}$ & AP$^b_\text{75}$ & AP$^m$ & AP$^m_\text{50}$ & AP$^m_\text{75}$ \\
        
        \midrule
        \multicolumn{9}{l}{\textbf{\textit{Mask R-CNN $\ \ 3\times\!$ Schedule}}} \\
        \midrule
    
        InternImage-T~\cite{internimage}
        & C & 270G & 49.1 & 70.4 & 54.1 & 43.7 & 67.3 & 47.3 \\
        CSWin-T~\cite{cswin}
        & T & 279G & 49.0 & 70.7 & 53.7 & 43.6 & 67.9 & 46.6 \\
        DAT-T++~\cite{dat++}
        & T & 301G & 50.5 & 71.9 & 55.7 & 45.1 & 69.2 & 48.7 \\
        Vmamba-T~\cite{vmamba}
        & M & 270G & 48.9 & 70.6 & 53.6 & 43.7 & 67.7 & 46.8 \\
        LocalVMamba-T~\cite{localmamba}
        & M & 291G & 48.7 & 70.1 & 53.0 & 43.4 & 67.0 & 46.4 \\
        PolaFormer-T~\cite{flatten}
        & L & 268G & 47.0 & 68.9 & 51.5 & 42.3 & 66.0 & 45.8 \\
        MILA-T~\cite{demystify_mamba}
        & L & 255G & 48.8 & 71.0 & 53.6 & 43.8 & 68.0 & 46.8 \\
        \rowcolor{lightgray!20} H-ViT$^3$-T
        &   & 271G & 48.9 & 71.0 & 53.4 & 44.0 & 68.0 & 47.5 \\

        \midrule
        InternImage-S~\cite{internimage}
        & C & 340G & 49.7 & 71.1 & 54.5 & 44.5 & 68.5 & 47.8 \\
        QFormer$_h$-S~\cite{quadrangle} 
        & T & -    & 49.5 & 71.2 & 54.2 & 44.2 & 68.3 & 47.6 \\
        CSWin-S~\cite{cswin}
        & T & 342G & 50.0 & 71.3 & 54.7 & 44.5 & 68.4 & 47.7 \\
        DAT-S++~\cite{dat++}
        & T & 430G & 51.2 & 72.6 & 56.3 & 45.7 & 69.9 & 49.7 \\
        Vmamba-S~\cite{vmamba}
        & M & 384G & 49.9 & 70.9 & 54.7 & 44.2 & 68.2 & 47.7 \\
        MILA-S~\cite{demystify_mamba}
        & L & 319G & 50.5 & 71.8 & 55.2 & 44.9 & 69.1 & 48.2 \\
        \rowcolor{lightgray!20} H-ViT$^3$-S
        &   & 349G & 50.5 & 72.0 & 55.5 & 45.0 & 69.1 & 48.8 \\

        \midrule
        InternImage-B~\cite{internimage}
        & C & 501G & 50.3 & 71.4 & 55.3 & 44.8 & 68.7 & 48.0 \\
        SWin-B~\cite{swin}
        & T & 526G & 48.6 & 70.0 & 53.4 & 43.3 & 67.1 & 46.7 \\
        CSWin-B~\cite{cswin}
        & T & 526G & 50.8 & 72.1 & 55.8 & 44.9 & 69.1 & 48.3 \\
        \rowcolor{lightgray!20} H-ViT$^3$-B
        &   & 510G & 51.0 & 72.1 & 55.9 & 45.3 & 69.3 & 49.1 \\
        
        \bottomrule
    \end{tabular}}
    \end{minipage}
    \vskip -0.2cm
    \caption{Results on COCO dataset. C, T, M, L represent ConvNet, Transformer, Mamba, and Linear attention, respectively. The FLOPs are computed with an input resolution of 1280$\times$800.}
    \label{tab:det}
    \vskip -0.5cm
\end{table*}

\vskip 0.1cm
\noindent
\textbf{Insight 6: convolutional architectures are particularly appropriate as inner models for visual tasks.}
Convolutional operations have long been a cornerstone of visual models~\cite{resnet, densenet, mobilenetv2} before the widespread adoption of Transformers~\cite{vit}. Owing to the flexibility of TTT framework, the inner model $\mathcal{F}_W(\cdot)$ can be implemented as a compact convolutional network instead of being restricted to modules based on linear layers (e.g., MLP or GLU) in prior work~\cite{ttt, lact, ttt_video}. As a showcase, we evaluate two simple designs: a standard $3\!\times\!3$ convolution, and a lightweight $3\!\times\!3$ depthwise convolution~\cite{mobilenet}. Results in Table \ref{tab:inner_model} show that both variants yield strong accuracy improvements. We argue that using a convolutional inner module provides a natural and elegant integration of global and local information. Recall that TTT compresses the global context $K,V$ into the parameters of $\mathcal{F}_W(\cdot)$. When $\mathcal{F}_W(\cdot)$ is implemented as a convolution, global information is effectively encoded into the local convolution kernel weights. Consequently, computing the output $O\!=\!\mathcal{F}_W(Q)$ realizes both global (via the updated kernel weights) and local (via the convolution receptive field) interactions, thus delivering a natural integration of global and local features and performance gains.

\noindent
\textbf{Remark 6.}
As discussed in \cref{sec:preliminaries}, TTT fits the inner model using a key–value dataset $\mathcal{D}=\{(K_i,V_i)|i=1,\cdots,N\}$. When the inner model is convolutional, we actually generalize this definition to $\mathcal{D}=\{(K^\mathrm{loc}_i,V_i)|i=1,\cdots,N\}$, where $K^\mathrm{loc}_i$ represents the local neighborhood of $K_i$. For instance, with a $3\!\times\!3$ convolution inner model, TTT learns pairs $(K^{3\times3}_i,V_i)$, where $K^{3\times3}_i$ represents the $3\!\times\!3\!=\!9$ key tokens in the local window centered at $K_i$.

\section{ViT$^3$: A Test-Time Training Architecture}

In \cref{sec:method}, we present a controlled study of visual TTT, distilling six practical insights that inform effective designs and potential future works. Guided by these findings, in this section we introduce Vision Test-Time Training (ViT$^3$) model, a pure TTT \textit{baseline} with linear complexity for benchmarking TTT methods on diverse visual tasks.

Specifically, for inner training, we use a single epoch of full-batch gradient descent with learning rate 1.0, optimizing a dot-product loss. As shown in \cref{sec:inner_training}, this simple configuration is efficient and effective for visual TTT. The inner model comprises two useful modules we identified in \cref{sec:inner_model}: a simplified gated linear unit $\mathcal{F}_1\!=\!\mathrm{FC}(x)\odot\mathrm{SiLU}(\mathrm{FC}(x))$ and a depthwise convolution $\mathcal{F}_2\!=\!\mathrm{DWConv}(x)$. The gated linear unit doubles the capacity of a naive $d\!\times\!d$ linear state while remaining easy to optimize, whereas the depthwise convolution offers a natural integration of local and global information. Within each TTT block, we use $\mathcal{F}_2$ in a single attention head and instantiate the remaining heads with $\mathcal{F}_1$. The resulting design is a drop-in replacement for standard attention blocks and can be integrated into various vision Transformer backbones. We practically build two model families, ViT$^3$ (non-hierarchical) and H-ViT$^3$ (hierarchical 4-stage), following the design philosophies of classical vision Transformers~\cite{vit, deit} and modern 4-stage architectures~\cite{swin,pvt}, respectively. We further adapt our approach to diffusion image Transformers (DiT)~\cite{dit} for generative tasks, building DiT$^3$ models. Detailed model architectures are shown in the Appendix.

\subsection{Image Classification}

The ImageNet-1K~\cite{imagenet} dataset contains 1.28M training images and 50K validation images across 1,000 classes. To ensure a fair comparison with prior work, we follow the training protocol used in Swin Transformer~\cite{swin}. Concretely, all models are trained from scratch for 300 epochs with the AdamW optimizer~\cite{adamw}, a cosine learning-rate decay and a linear warm-up for the first 20 epochs. We use a weight decay of 0.05. The total batch size is 4096 and initial learning rate is set to $4\times{10}^{-3}$. Standard augmentation and regularization methods are applied, including RandAugment~\cite{randaugment}, Mixup~\cite{mixup}, CutMix~\cite{cutmix}, and random erasing~\cite{random_erasing}. We also report results with MESA~\cite{mesa} training strategy.

The results are presented in \cref{tab:cls1} and \cref{tab:cls2}. We observe that ViT$^3$ and H-ViT$^3$ models consistently outperform various linear attention and Mamba variants, while remaining competitive with state-of-the-art vision Transformer models. For example, H-ViT$^3$-S attains higher performance than the larger SOFT-L++~\cite{soft_plus} and VMamba-B~\cite{vmamba}, despite using roughly half of parameters and FLOPs. These results validate the advantages of the TTT paradigm over prior linear-complexity designs and highlight its potential for efficient, scalable $\mathcal{O}(N)$ visual sequence modeling.

\subsection{Object Detection}

The COCO \cite{coco} dataset is a widely adopted benchmark for object detection and instance segmentation. We employ the standard 1x and 3x Mask R-CNN~\cite{mrcn} training schedules, presenting the results in \cref{tab:det}. With high-resolution inputs, token sequences in object detection are typically much longer than in image classification~\cite{mambaout}, resulting in $N\!\gg\!d$. Under these conditions, existing linear-complexity designs such as VMamba~\cite{vmamba} and SOFT++~\cite{soft_plus} tend to be constrained by limited state capacity and therefore frequently underperform. In contrast, with nonlinear inner modules and the online learning procedure, H-ViT$^{3}$ attains stronger global modeling: it matches or surpasses state-of-the-art linear-complexity methods and greatly narrows the performance gap to highly optimized vision Transformers.

\subsection{Semantic Segmentation}

The ADE20K~\cite{ade20k} is a well-established benchmark for semantic segmentation. We use UPerNet~\cite{upernet} as the framework. As shown in \cref{tab:seg}, similar to the trend in detection task, H-ViT$^{3}$ establishes a strong linear-complexity baseline for semantic segmentation and consistently outperforms VMamba~\cite{vmamba}, VVT~\cite{vvt} and SOFT++~\cite{soft_plus}. Nevertheless, H-ViT$^{3}$ remains inferior to highly optimized vision Transformers such as TransNeXt~\cite{transnext}. We argue that designing deeper, more expressive inner modules is a promising direction to further boost capacity and to build TTT models comparable with state-of-the-art Transformers.

\begin{table}[t]

    \centering
    \footnotesize
    \setlength{\tabcolsep}{1.5mm}{
    \renewcommand\arraystretch{0.9}
    \begin{tabular}{l c c c c}
        \toprule
        Backbone & Type & \#Params & FLOPs & mIoU \\

        \midrule
        ConvNeXt-T~\cite{convnext}
        & ConvNet       & 60M & 939G & 46.0 \\
        Focal-T~\cite{focal}
        & Transformer   & 62M & 998G & 45.8 \\
        FasterViT-2~\cite{nat}
        & Transformer   & -   & 974G & 47.2 \\
        VMamba-T~\cite{vmamba}
        & Mamba         & 62M & 949G & 47.9 \\
        SOFT-T++~\cite{soft_plus}
        & Linear        & 60M & 948G & 46.5 \\
        VVT-S~\cite{vvt}
        & Linear        & 56M & 960G & 46.8 \\
        \rowcolor{lightgray!20} H-ViT$^3$-T
        & TTT           & 58M & 946G & 48.0 \\

        \midrule
        ConvNeXt-S~\cite{convnext}
        & ConvNet       & 82M & 1027G & 48.7 \\
        CSwin-S~\cite{cswin}
        & Transformer   & 65M & 1027G & 50.4 \\
        TransNeXt-S~\cite{transnext}
        & Transformer   & 80M & 1089G & 52.2 \\
        LocalVMamba-S~\cite{localmamba}
        & Mamba         & 81M & 1095G & 50.0 \\
        VVT-M~\cite{vvt}
        & Linear        & 78M & 1040G & 48.1 \\
        SOFT-S++~\cite{soft_plus}
        & Linear        & 81M & 1040G & 48.9 \\
        \rowcolor{lightgray!20} H-ViT$^3$-S
        & TTT           & 84M & 1026G & 50.2 \\

        \midrule
        ConvNeXt-B~\cite{convnext}
        & ConvNet       & 122M & 1170G & 49.1 \\
        CSwin-B~\cite{cswin}
        & Transformer   & 109M & 1222G & 51.1 \\
        TransNeXt-B~\cite{transnext}
        & Transformer   & 121M & 1268G & 53.0 \\
        VMamba-B~\cite{vmamba}
        & Mamba         & 122M & 1170G & 51.0 \\
        VVT-L~\cite{vvt}
        & Linear        & 92M  & 1068G & 48.8 \\
        SOFT-B++~\cite{soft_plus}
        & Linear        & 121M & 1204G & 49.2 \\
        \rowcolor{lightgray!20} H-ViT$^3$-B
        & TTT           & 124M & 1195G & 51.7 \\
        \bottomrule
    \end{tabular}}
    \vskip -0.2cm
    \caption{Results of semantic segmentation. FLOPs are calculated with an input resolution of 512$\times$2048.}
    \label{tab:seg}
    \vskip -0.2cm
\end{table}

\subsection{Image Generation}

We further benchmark TTT method on the class-conditional image generation task using the ImageNet-1K dataset. Specifically, we replace the Softmax attention module in DiT~\cite{dit} with our ViT$^3$ block, yielding the DiT$^3$ model family. Following standard protocol, we report FID on 50 000 validation samples (FID-50K) at $256^2$ resolution. \cref{tab:generation} shows that DiT$^3$ consistently improves DiT in various settings without extra tuning. These results validate the effectiveness of ViT$^3$ for image generation, and establish strong TTT baselines for linear-complexity approaches.

\begin{table}[t]

    \centering
    \footnotesize
    \setlength{\tabcolsep}{1.4mm}{
    \renewcommand\arraystretch{0.9}
    \begin{tabular}{l c c c c c c}
        \toprule
        Model & \#Params & FLOPs & FID$\downarrow$ & IS$\uparrow$ & Prec.$\uparrow$& Rec.$\uparrow$ \\ 
        
        \midrule
        DiT-S/8~\cite{dit}
        & 33M & 0.36G & 153.60 & -     & -    & - \\ 
        \rowcolor{lightgray!20} DiT$^3$-S/8 
        & 35M & 0.40G & 143.49 & 8.34  & 0.15 & 0.13  \\ 
        DiT-S/4~\cite{dit}
        & 33M & 1.41G & 100.41 & -     & -    & -  \\ 
        \rowcolor{lightgray!20} DiT$^3$-S/4
        & 35M & 1.57G & 93.77  & 14.42 & 0.27 & 0.40  \\ 
        DiT-S/2~\cite{dit}
        & 33M & 6.06G & 68.40  & -     & -    & - \\ 
        \rowcolor{lightgray!20} DiT$^3$-S/2 
        & 35M & 6.23G & 62.65  & 21.59 & 0.39 & 0.57    \\ 
        
        \midrule
        DiT-B/8~\cite{dit}
        & 131M & 1.42G  & 122.74 & -     &-     & -  \\ 
        \rowcolor{lightgray!20} DiT$^3$-B/8 
        & 135M & 1.51G  & 120.41 & 10.94 & 0.20 & 0.26  \\ 
        DiT-B/4~\cite{dit}
        & 130M & 5.56G  & 68.38  & -     &-     & - \\ 
        \rowcolor{lightgray!20} DiT$^3$-B/4
        & 134M & 5.88G  & 65.25  & 22.29 & 0.37 & 0.55 \\ 
        DiT-B/2~\cite{dit}
        & 130M & 23.01G & 43.47  & -     & -    &- \\ 
        \rowcolor{lightgray!20} DiT$^3$-B/2 
        & 134M & 23.35G & 39.31  & 36.99 &0.51  & 0.62  \\ 
        
        \bottomrule
    \end{tabular}}
    \vskip -0.2cm
    \caption{Results of class-conditional image generation.}
    \label{tab:generation}
    \vskip -0.2cm
    
\end{table}

\begin{figure}[t]
    \centering
    \includegraphics[width=\linewidth]{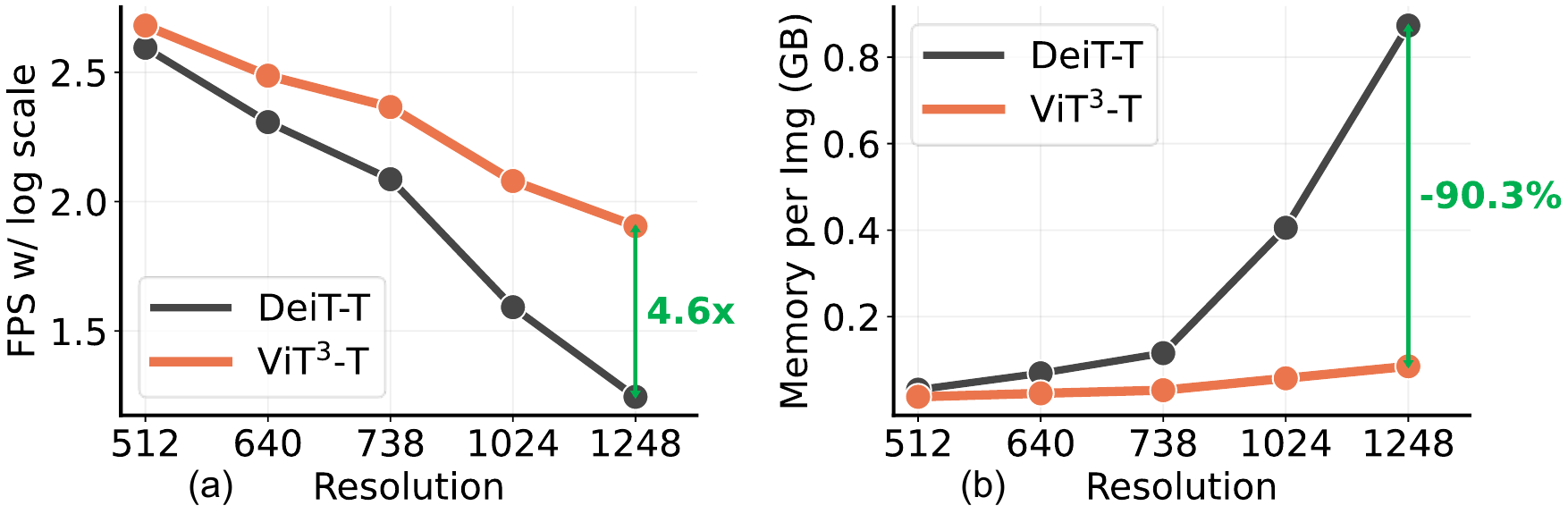}
    \vskip -0.1in
    \caption{Comparisons between DeiT and ViT$^3$ in (a) FPS on RTX3090, and (b) per image GPU memory usage. }
    \label{fig:fps_memory}
    \vskip -0.2cm
\end{figure}

\subsection{Efficiency Analysis}

The above evaluations primarily examine the expressiveness of TTT method. Here we would like to highlight its efficiency advantages over the widely adopted Softmax attention. \cref{fig:fps_memory} compares the FPS and GPU memory usage of ViT$^3$-T and DeiT-T. Owing to its linear time and memory complexity, ViT$^3$ scales more efficiently in both throughput and memory consumption as image resolution increases. At $1248^2$ resolution (i.e., 6,084 tokens per image), ViT$^3$-T attains a 4.6× speedup over DeiT-T and reduces GPU memory consumption by 90.3\%, demonstrating its efficiency.


\section{Conclusion}

Designing efficient yet expressive sequence-modeling paradigms remains a key pursuit in computer vision. In this paper, we systematically study the design space of Test-Time Training (TTT), a new promising approach for building scalable models with linear complexity. Through a series of experiments, we distill our observations into six practical insights, shedding some light on the design principles for effective visual TTT and possible future directions. Our findings and analyses culminated in the Vision Test-Time Training (ViT$^3$) model, a pure TTT architecture for visual sequence modeling. ViT$^3$ achieves competitive results across multiple tasks, serving as a strong baseline for linear-complexity methods. We hope our findings and analyses can stimulate further research on visual TTT models.

\section*{Acknowledgements}
This work is supported in part by the National Key R\&D Program of China under Grant 2024YFB4708200, the National Natural Science Foundation of China under Grants U24B20173 and U2541227, and the Scientific Research Innovation Capability Support Project for Young Faculty under Grant ZYGXQNJSKYCXNLZCXM-I20.

{
    \small
    \bibliographystyle{ieeenat_fullname}
    \bibliography{main}
}

\clearpage
\setcounter{page}{1}
\maketitlesupplementary

\section{Contribution and Limitation}

In this paper, we systematically study the design space of Test-Time Training (TTT), shedding some light on the design principles for effective visual TTT and possible future directions. Our main contributions are as follows:

\begin{itemize}

    \item We present a systematic empirical study of Test-Time Training designs for vision, covering inner training regimes (loss function, learning rate, batch size, epochs) and inner model design (architecture and model size).
    
    \item We offer six practical insights for building effective yet efficient TTT module, providing detailed analyses of the TTT design space. Our analyses also reveal several valuable future research directions for TTT models.

    \item We build the Vision Test-Time Training (ViT$^3$) model, a simple TTT architecture that implements these insights. With $\mathcal{O}(N)$ complexity, ViT$^3$ achieves competitive results across image classification, image generation, object detection, and semantic segmentation, serving as a strong baseline for future research on visual TTT models.
\end{itemize}

However, we note that there are other design choices that we have not covered (e.g., inner optimizer, inner data augmentation, Transformer inner model, etc.), and this paper is not exhaustive. Exploring these axes is left to future work.

\section{Inner Training Loss}
Consider a mini-batch of target value tokens and model predictions $V_{\mathcal{B}},\hat{V}_{\mathcal{B}}\in\mathbb{R}^{B\times d}$, where $B$ denotes the batch size. We denote the $i$-th token (row) by $V_i,\hat{V}_i\in\mathbb{R}^{1\times d}$. 

For each loss function considered in Tab.~1 of the main paper, we provide the explicit formula and compute the mixed second derivative $\frac{\partial^2\mathcal{L}}{\partial V_{ij}\partial \hat{V}_{ij}}$.

\textbf{(1) Dot Product Loss.}
\begin{equation}
    \mathcal{L}=-\frac{1}{B\sqrt{d}} \sum_{i=1}^{B} \hat{V}_i V_i^\top.
\end{equation}

The mixed second derivative is:
\begin{equation}
    \begin{split}
        \frac{\partial^2\mathcal{L}}{\partial V_{ij}\partial \hat{V}_{ij}} &= \frac{\partial}{\partial V_{ij}} \left( \frac{\partial \mathcal{L}}{\partial \hat{V}_{ij}} \right) \\
        &= \frac{\partial}{\partial V_{ij}} \left( -\frac{1}{B\sqrt{d}} V_{ij} \right) \\
        &= -\frac{1}{B\sqrt{d}}.
    \end{split}
\end{equation}

\textbf{(2) MSE (L2) Loss.}
\begin{equation}
    \mathcal{L}=\frac{1}{2B\sqrt{d}} \sum_{i=1}^{B} (\hat{V}_i-V_i) (\hat{V}_i-V_i)^\top.
\end{equation}

The mixed second derivative is:
\begin{equation}
    \begin{split}
        \frac{\partial^2\mathcal{L}}{\partial V_{ij}\partial \hat{V}_{ij}} &= \frac{\partial}{\partial V_{ij}} \left( \frac{\partial \mathcal{L}}{\partial \hat{V}_{ij}} \right) \\
        &= \frac{\partial}{\partial V_{ij}} \left( \frac{1}{B\sqrt{d}} (\hat{V}_{ij} - V_{ij}) \right) \\
        &= -\frac{1}{B\sqrt{d}}.
    \end{split}
\end{equation}

\textbf{(3) RMSE Loss.}
\begin{equation}
    \mathcal{L}=\sqrt{\frac{1}{B\sqrt{d}} \sum_{i=1}^{B} (\hat{V}_i-V_i) (\hat{V}_i-V_i)^\top}.
\end{equation}

The mixed second derivative is:
\begin{equation}
    \begin{split}
        \frac{\partial^2\mathcal{L}}{\partial V_{ij}\partial \hat{V}_{ij}} &= \frac{\partial}{\partial V_{ij}} \left( \frac{\partial \mathcal{L}}{\partial \hat{V}_{ij}} \right) \\
        &= \frac{\partial}{\partial V_{ij}} \left( \frac{1}{B\sqrt{d} \sqrt{S}} (\hat{V}_{ij} - V_{ij}) \right) \\
        &= -\frac{1}{B\sqrt{d} \sqrt{S}} + \frac{1}{B^2 d S^{3/2}} (\hat{V}_{ij} - V_{ij})^2, \\
        S&=\frac{1}{B\sqrt{d}} \sum_{i=1}^{B} (\hat{V}_i-V_i) (\hat{V}_i-V_i)^\top.
    \end{split}
\end{equation}

\textbf{(4) MAE (L1) Loss.}
\begin{equation}
    \mathcal{L}=\frac{1}{B\sqrt{d}} \sum_{i=1}^{B} \|\hat{V}_i-V_i\|_1,
\end{equation}

The mixed second derivative is:
\begin{equation}
    \begin{split}
        \frac{\partial^2\mathcal{L}}{\partial V_{ij}\partial \hat{V}_{ij}} &= \frac{\partial}{\partial V_{ij}} \left( \frac{\partial \mathcal{L}}{\partial \hat{V}_{ij}} \right) \\
        &= \frac{\partial}{\partial V_{ij}} \left( \frac{1}{B\sqrt{d}} \operatorname{sign}(\hat{V}_{ij} - V_{ij}) \right) \\
        &= 0 \quad \text{if } \hat{V}_{ij} \neq V_{ij}.
    \end{split}
\end{equation}

\textbf{(5) Smooth L1 loss.}
\begin{equation}
    \begin{split}
        \mathcal{L}&=\frac{1}{B\sqrt{d}} \sum_{i=1}^{B} \sum_{j=1}^{d} \ell(\hat{V}_{ij}-V_{ij}), \\
        \ell(x)&=\begin{cases} \frac{1}{2}x^2 & \text{if } |x| < 1 \\ |x| - \frac{1}{2} & \text{otherwise} \end{cases}.
    \end{split}
\end{equation}

The mixed second derivative is:
\begin{equation}
    \begin{split}
        \frac{\partial^2\mathcal{L}}{\partial V_{ij}\partial \hat{V}_{ij}} &= \frac{\partial}{\partial V_{ij}} \left( \frac{\partial \mathcal{L}}{\partial \hat{V}_{ij}} \right) \\
        &= \frac{\partial}{\partial V_{ij}} \left( \frac{1}{B\sqrt{d}} \ell'(\hat{V}_{ij} - V_{ij}) \right) \\
        &= -\frac{1}{B\sqrt{d}} \ell''(\hat{V}_{ij} - V_{ij}) \\
        &= -\frac{1}{B\sqrt{d}} \times \begin{cases} 1 & \text{if } |\hat{V}_{ij} - V_{ij}| < 1 \\ 0 & \text{if } |\hat{V}_{ij} - V_{ij}| > 1 \end{cases}.
    \end{split}
\end{equation}

Notably, the $1/\sqrt{d}$ scaling used above is consistent with the scaled dot product attention convention~\cite{attention}. As analyzed in the main paper, losses with vanishing (or piecewise-vanishing) mixed second derivatives — in particular MAE (almost everywhere zero) and Smooth L1 in its linear region — hinder the learning of outer model parameter $W_V$ matrix and therefore leads to lower performance.

\section{Model Architecture}

As discussed in the main paper, we present a plug-in visual TTT block based on our findings. Specifically, for inner training, we use a single epoch of full-batch gradient descent with learning rate 1.0, optimizing a dot-product loss. The inner model comprises a simplified gated linear unit $\mathcal{F}_1\!=\!\mathrm{FC}(x)\odot\mathrm{SiLU}(\mathrm{FC}(x))$ and a depthwise convolution $\mathcal{F}_2\!=\!\mathrm{DWConv}(x)$. The gated linear unit doubles the capacity of a naive $d\!\times\!d$ linear state while remaining easy to optimize, whereas the depthwise convolution offers a natural integration of local and global information. Within each TTT block, we use $\mathcal{F}_2$ in a single attention head and instantiate the remaining heads with $\mathcal{F}_1$. 

We build two model families with this TTT block, ViT$^3$ (non-hierarchical) and H-ViT$^3$ (hierarchical 4-stage), and adapt our approach to diffusion image Transformers (DiT)~\cite{dit} for generative tasks. The architectures are provided in \cref{tab:vit3_arch}, \cref{tab:h_vit3_arch}, and \cref{tab:dit3_arch}. To introduce positional information, we employ conditional positional encodings~\cite{cpe}, which is widely adopted by modern vision Transformers~\cite{cmt, biformer, dat++}, linear attention~\cite{demystify_mamba} and Mamba models~\cite{vim, vmamba, localmamba}. Since our method benefits from $\mathcal{O}(N)$ complexity, we directly process the high-resolution feature map with a global receptive field.

\begin{table}[t]
    \centering
    \footnotesize
    \setlength{\tabcolsep}{1.6mm}{
    \renewcommand\arraystretch{1.2}
    \begin{tabular}{c|c|c|c}
        \toprule
                                & ViT$^3$-T     & ViT$^3$-S     & ViT$^3$-B  \\

        \midrule
        \multirow{2}{*}{Backbone} 
        & Patch$\ \downarrow\!16$ & Patch$\ \downarrow\!16$ & Patch$\ \downarrow\!16$ \\
        & ${\rm B}(192, 6)\!\times\!12$ & ${\rm B}(384, 6)\!\times\!12$ & ${\rm B}(768, 12)\!\times\!12$ \\

        \midrule
        Classifier & \multicolumn{3}{c}{Global Average Pooling, Linear} \\
        
        \bottomrule
    \end{tabular}}
    \caption{Architectures of ViT$^3$ model series. Patch ``$\downarrow\!n$'' indicates the patch size is $n$. ``${\rm B}(C, H)$'' represents one building block with embedding dimension $C$ and $H$ attention heads.}
    \label{tab:vit3_arch}
\end{table}

\begin{table}[t]
    \centering
    \footnotesize
    \setlength{\tabcolsep}{0.8mm}{
    \renewcommand\arraystretch{1.2}
    \begin{tabular}{c|c|c|c|c}
        \toprule
                                & Size      & H-ViT$^3$-T     & H-ViT$^3$-S     & H-ViT$^3$-B  \\

        \midrule
        \multirow{2}{*}{Stage1} & \multirow{2}{*}{$\frac{H}{4}\!\!\times\!\!\frac{W}{4}$}
        & Stem$\ \downarrow\!4$ & Stem$\ \downarrow\!4$ & Stem$\ \downarrow\!4$ \\
        & & ${\rm B}(64, 2)\!\times\!1$ & ${\rm B}(64, 2)\!\times\!2$ & ${\rm B}(96, 3)\!\times\!2$ \\

        \midrule
        \multirow{2}{*}{Stage2} & \multirow{2}{*}{$\frac{H}{8}\!\!\times\!\!\frac{W}{8}$}
        & Down$\ \downarrow\!2$ & Down$\ \downarrow\!2$ & Down$\ \downarrow\!2$ \\
        & & ${\rm B}(128, 4)\!\times\!3$ & ${\rm B}(128, 4)\!\times\!6$ & ${\rm B}(192, 6)\!\times\!6$ \\

        \midrule
        \multirow{2}{*}{Stage3} & \multirow{2}{*}{$\frac{H}{16}\!\!\times\!\!\frac{W}{16}$}
        & Down$\ \downarrow\!2$ & Down$\ \downarrow\!2$ & Down$\ \downarrow\!2$ \\
        & & ${\rm B}(320, 10)\!\times\!9$ & ${\rm B}(320, 10)\!\times\!18$ & ${\rm B}(448, 14)\!\times\!18$ \\

        \midrule
        \multirow{2}{*}{Stage4} & \multirow{2}{*}{$\frac{H}{32}\!\!\times\!\!\frac{W}{32}$}
        & Down$\ \downarrow\!2$ & Down$\ \downarrow\!2$ & Down$\ \downarrow\!2$ \\
        & & ${\rm B}(512, 16)\!\times\!4$ & ${\rm B}(512, 16)\!\times\!8$ & ${\rm B}(640, 20)\!\times\!8$ \\

        \midrule
        Classifier & & \multicolumn{3}{c}{Global Average Pooling, Linear} \\
        
        \bottomrule
    \end{tabular}}
    \caption{Architectures of H-ViT$^3$ model series. ``$\downarrow\!n$'' indicates the downsampling ratio is $n$. ``${\rm B}(C, H)$'' represents one building block with embedding dimension $C$ and $H$ attention heads.}
    \label{tab:h_vit3_arch}
\end{table}

\begin{table}[t]
    \centering
    \footnotesize
    \setlength{\tabcolsep}{1.6mm}{
    \renewcommand\arraystretch{1.2}
    \begin{tabular}{c|c|c|c}
        \toprule
                                & DiT$^3$-S/8     & DiT$^3$-S/4     & DiT$^3$-S/2  \\

        \midrule
        \multirow{2}{*}{Backbone} 
        & Patch$\ \downarrow\!8$ & Patch$\ \downarrow\!4$ & Patch$\ \downarrow\!2$ \\
        & ${\rm B}(384, 6)\!\times\!12$ & ${\rm B}(384, 6)\!\times\!12$ & ${\rm B}(384, 6)\!\times\!12$ \\

        \bottomrule
        \toprule
                                & DiT$^3$-B/8     & DiT$^3$-B/4     & DiT$^3$-B/2  \\

        \midrule
        \multirow{2}{*}{Backbone} 
        & Patch$\ \downarrow\!8$ & Patch$\ \downarrow\!4$ & Patch$\ \downarrow\!2$ \\
        & ${\rm B}(768, 12)\!\times\!12$ & ${\rm B}(768, 12)\!\times\!12$ & ${\rm B}(768, 12)\!\times\!12$ \\
        
        \bottomrule
    \end{tabular}}
    \caption{Architectures of DiT$^3$ model series. Patch ``$\downarrow\!n$'' indicates the patch size is $n$. ``${\rm B}(C, H)$'' represents one building block with embedding dimension $C$ and $H$ attention heads.}
    \label{tab:dit3_arch}
\end{table}


\end{document}